\documentclass{article}
\usepackage{spconf,amsmath,graphicx}
\usepackage{amssymb}
\usepackage{booktabs}
\usepackage{lipsum}
\usepackage{bm}
\usepackage{stmaryrd}
\usepackage{dsfont}
\usepackage{multirow}
\usepackage{algorithm}
\usepackage{algpseudocode}
\usepackage[implicit=false]{hyperref}

\usepackage[capitalize]{cleveref}
\crefname{section}{Sec.}{Secs.}
\Crefname{section}{Section}{Sections} 
\Crefname{table}{Table}{Tables}
\crefname{table}{Tab.}{Tabs.}
\Crefname{equation}{Equation}{Equations}

\title{EXTRACTING EFFECTIVE SUBNETWORKS WITH GUMBEL-SOFTMAX}
\name{Robin Dupont$^{\star \dagger}$ \qquad Mohammed Amine Alaoui$^{\dagger}$ \qquad Hichem Sahbi$^{\star}$ \qquad Alice Lebois$^{\dagger}$}

\address{$^{\star}$ Sorbonne Université, LIP6, Paris \\ 
    $^{\dagger}$ Netatmo, Boulogne-Billancourt}

\begin{document}
%
\maketitle
\begin{abstract}
Large and performant neural networks are often overparameterized and can be
drastically reduced in size and complexity thanks to pruning. Pruning is a group
of methods, which seeks to remove redundant or unnecessary weights or groups of
weights in a network. These techniques allow the creation of lightweight
networks, which are particularly critical in embedded or mobile applications. \\
In this paper, we devise an alternative pruning method that allows extracting
effective subnetworks from larger untrained ones. Our method is stochastic and
extracts subnetworks by exploring different topologies which are sampled using
Gumbel Softmax. The latter is also used to train probability distributions which
measure the relevance of weights in the sampled topologies. The resulting
subnetworks are further enhanced using a highly efficient rescaling mechanism
that reduces training time and improves performance. Extensive experiments
conducted on CIFAR show the outperformance of our subnetwork extraction method
against the related work.

\end{abstract}
  
\begin{keywords}
Lightweight networks, pruning, efficient computation, topology selection
\end{keywords}

  

  \section{Introduction}
  \label{sec:intro}

Deep neural networks are nowadays becoming mainstream in solving many image
processing tasks including visual category recognition. The success of these
models has been reached at the expense of an increase in their inference time,
memory consumption and energy footprint. With the era of intelligent embedded systems
(provided with limited energy and computational resources), a current trend is
to make these models {\it lightweight and frugal} while maintaining their high
accuracy.   Existing solutions in lightweight network design are targeted
toward creating small and efficient architectures from scratch
\cite{DBLP:conf/cvpr/HuangLMW18, DBLP:conf/cvpr/SandlerHZZC18,
DBLP:journals/corr/HowardZCKWWAA17, DBLP:conf/icml/TanL19} while others derive
highly compact yet effective neural networks from larger ones. These methods
predominantly include knowledge distillation
\cite{DBLP:journals/corr/HintonVD15, DBLP:conf/iclr/ZagoruykoK17,
DBLP:journals/corr/RomeroBKCGB14, DBLP:conf/aaai/MirzadehFLLMG20,
DBLP:conf/cvpr/ZhangXHL18, DBLP:conf/cvpr/AhnHDLD19} and pruning
\cite{DBLP:conf/nips/CunDS89, DBLP:conf/nips/HassibiS92,
DBLP:conf/nips/HanPTD15}. \\  \indent Pruning methods, either structured or
unstructured, are particularly successful, and seek to remove connections with
the least perceptible impact on classification accuracy. Structured pruning
consists in {\it jointly} removing groups of weights, entire channels or
subnetworks   \cite{DBLP:conf/iclr/0022KDSG17, DBLP:conf/iccv/LiuLSHYZ17},
whereas unstructured pruning aims at removing weights {\it individually}
\cite{DBLP:conf/nips/HanPTD15,DBLP:journals/corr/HanMD15}.  Unstructured
pruning has witnessed a recent surge in interest in the wake of the Lottery
Ticket Hypothesis  \cite{DBLP:conf/iclr/FrankleC19}; an empirical study in
\cite{DBLP:conf/iclr/FrankleC19} shows that large pretrained networks
encompass subnetworks, called \textit{Lottery Tickets}, whose
training with initial weights taken from the large networks yields
comparably accurate classifiers. Another study
\cite{DBLP:conf/iclr/LiuSZHD19} pushes that finding further and concludes that
only the topology of these subnetworks is actually important in order to reach
comparable performances.  In general, extracting an efficient subnetwork is
still an open problem and is computationally demanding as this amounts to full
training of large networks (till convergence) prior to their pruning. Existing
alternatives approach this problem using early pruning
\cite{DBLP:conf/iclr/LeeAT19,
DBLP:conf/iclr/WangZG20,DBLP:conf/nips/TanakaKYG20}, but still require to
train the weights. In contrast to these works, our
proposed solution in this paper identifies effective subnetworks by training
only their topology and without any weights tuning.\\  
\indent A theoretical analysis in
\cite{DBLP:conf/icml/MalachYSS20,DBLP:conf/nips/PensiaRNVP20,DBLP:conf/nips/OrseauHR20}
has established the sufficient conditions about the existence of efficient and
effective subnetworks in over-parameterized large networks, nonetheless, no
constructive proof has been provided in order to identify these subnetworks. In
this context, Zhou et al. \cite{DBLP:conf/nips/ZhouLLY19} proposed the first
attempt to extract efficient subnetworks using stochastic mask training. A
probability of selecting each weight is defined (as the sigmoid of a mask) and
trained using the Straight Through Estimator (STE)
\cite{DBLP:journals/corr/BengioLC13}. During training, weights are frozen and
only the masks are allowed to vary. However, the major drawback of this method
resides in the vanishing gradient of the sigmoid which makes mask training
numerically challenging. Ramanujan et al. \cite{DBLP:conf/cvpr/RamanujanWKFR20}
proposed another alternative, based on binarized saliency indicators learned
with STE, which selects the most prominent weights in the resulting subnetworks.
Nevertheless, since this method enfore the pruning rate \textit{a priori},
finding the pruning rate giving the higest performances has to be made through
a cumbersome and time-consuming binary search or grid-search. \\ 
\indent Considering the limitation of
the aforementioned related  work, we introduce in this paper a new stochastic
subnetwork selection method based on Gumbel Softmax. The latter allows sampling
subnetworks whose weights are the most relevant for classification. The proposed
contribution also relies on a new mask parametrization, dubbed as Arbitrarily
Shifted Log Parametrization (ASLP), that allows a better conditioning of the
gradient and thereby mitigates numerical instability during mask optimization.
Besides, when combining ASLP with a learned weight rescaling mechanism, training
is accelerated and the accuracy of the resulting subnetworks improves as shown
later in experiments.
  
\section{Proposed Method}

Let $f_\theta$ be a deep neural network whose weights defined as $\theta =
\left\{\bm{w}_1,\mathellipsis, \bm{w}_L \right\}$, with $L$ being its depth,
$\bm{w}_\ell \in \mathbb{R}^{d_{\ell} \times d_{\ell-1}}$ its
$\ell^\textrm{th}$ layer weights, and $d_\ell$ the dimension of $\ell$. The
output of a given layer $\ell$ is defined as 
\begin{equation}
  \label{eq:layer_eq}
  \mathbf{z}_{\ell} = g_\ell(\bm{w}_\ell \otimes \mathbf{z}_{\ell-1}),
\end{equation}
being  $g_\ell$ an activation function and $\otimes$ the usual matrix product.
Without a loss of generality, we omit the bias in the definition of
(\ref{eq:layer_eq}).

\subsection{Stochastic Weight Sampling}
\indent Given a network $f_\theta$, weight pruning consists in removing
connections in the graph of $f_\theta$. A node in this graph refers to a
neural unit while an edge corresponds  to a cross-layer connection. Pruning is
usually obtained by freezing and zeroing-out  a subset of weights in $\theta$,
and this is achieved  by multiplying $\bm{w}_\ell$ by a binary mask
$\bm{m}_\ell \in \{ 0,1 \}^{\text{dim}(\bm{w}_\ell)}$. The
binary entries of $\bm{m}_\ell$ are set depending on whether the underlying
layer connections are kept or removed, so \Cref{eq:layer_eq} becomes
\begin{equation}
  \label{eq:pruned_layer_eq}
  \mathbf{z}_{\ell} = g_\ell( (\bm{m}_\ell \odot \bm{w}_\ell ) \otimes \mathbf{z}_{\ell-1} ).
\end{equation}
Here $\odot$ stands for the element-wise matrix product. In this definition,
the masks $\{\bm{m}_\ell\}_\ell$ are stochastic and sampled from a Bernoulli
distribution.\\

\noindent\textbf{Straight Through Estimator.} Zhou et
al.~\cite{DBLP:conf/nips/ZhouLLY19} consider a Bernoulli parametrization of
$\{\bm{m}_\ell\}_\ell$ in order to sample masks in \Cref{eq:pruned_layer_eq}.
However, due to sampling which is not a differentiable operation, optimizing directly
$\{{\bm{m}_\ell}\}_\ell$ is not possible. Existing
solutions, including \cite{DBLP:conf/nips/ZhouLLY19}, rely on the Straight
Trough Estimator (STE), already described
in~\cite{DBLP:journals/corr/BengioLC13}.  The definition of
$\{\bm{m}_\ell\}_\ell$ is instead based on another {\it latent}
parametrization $\{\bm{\hat{m}}_\ell\}_\ell$, detailed subsequently, and
obtained by applying a sigmoid function $\sigma(.)$ to $\bm{\hat{m}}_\ell$.
This allows optimizing $\bm{\hat{m}}_\ell$  using gradient descent while
considering the following surrogate of  \Cref{eq:pruned_layer_eq} 
\begin{equation}
  \label{eq:pruned_layer_eq2}
  \mathbf{z}_{\ell} = g_\ell( ( \sigma(\bm{\hat{m}}_\ell) \odot \bm{w}_\ell ) \otimes \mathbf{z}_{\ell-1} ).
\end{equation}

\noindent Authors in  \cite{DBLP:conf/nips/ZhouLLY19} use the STE in
order to back-propagate the gradient and to update the parameters of the
Bernoulli distribution $\bm{\hat{m}}_\ell$ with gradient descent.\\

\noindent\textbf{Gumbel-Softmax.} In what follows, we consider an alternative
STE based on Gumbel Softmax (GS)~\cite{DBLP:conf/iclr/JangGP17}. The proposed
method, dubbed as Straight Through Gumbel Softmax (STGS), is based (i) on a
variant of GS, and also (ii) on the argmax operator which allows sampling from
a categorical distribution, as the limit of GS (i.e., when its softmax temperature
approaches zero).  Let $z$ be a categorical random variable, associated with $n$
class probability distribution $\mathcal{P} = [\pi_1,\dots,\pi_n]$. The Gumbel
Softmax estimator (i) takes a vector of log-probabilities $\log(\mathcal{P})
=[\log(\pi_1),\dots, \log(\pi_n)]$ as an input, (ii) disrupts the latter with
a random additive noise sampled from the Gumbel distribution, and (iii) takes
the argmax, yielding a categorical variable. More formally, following
\cite{DBLP:conf/iclr/JangGP17}, the value $q$ of our categorical variable $z$
is obtained as 
\begin{equation}
  \label{eq:gumbel-softmax-argmax}
  q = \underset{k}{ \text{argmax}} \ [ \log(\pi_k)+g_k ],
\end{equation}
with $g_k$ being i.i.d sampled from  the Gumbel distribution.\\
\noindent In what follows, and unless stated otherwise, we omit $\ell$ from
$\bm{w}_\ell$ and we write it for short as $\bm{w}$. Let $\bm{w}_{ij}$ be the
weight associated to the i-th and j-th neurons  respectively belonging to
layers $\ell-1$ and $\ell$; we define a two-class categorical
distribution $\mathcal{P}_{ij}$ on $\{0,1\}$ as
$\mathcal{P}_{ij}(z=1)=\pi_1^{ij}$, and $\mathcal{P}_{ij}(z=0)=\pi_2^{ij}$
with $\pi_1^{ij}=p_{ij}$, $\pi_2^{ij}=1-p_{ij}$ and $p_{ij}$ being the
probability to keep the underlying connection. In other words, keeping the
weight $\bm{w}_{ij}$ (or not) in the sampled topology is a Bernoulli trial
with a probability $p_{ij}$. Considering
\Cref{eq:gumbel-softmax-argmax}, a binary mask  $\bm{m}_{ij}$ is defined as
$1_{\{q_{ij}=1\}}$, $1_{\{\}}$ being the indicator function and $q_{ij} =
{\text{argmax}_{k \in \{1,2\}}}\big[\log(\pi_k^{ij})+g_k^{ij}\big]$.
\noindent Thanks to STGS, it becomes possible to learn $p_{ij}$ for each
weight through stochastic gradient descent (SGD). However, optimizing $p_{ij}$
(with SGD) raises a major issue as $p_{ij}$ may not be appropriately bounded
and thereby $\log(p_{ij})$ and $\log(1-p_{ij})$ would also be undefined.  On
another hand,  solving constrained SGD, besides being computationally
expensive and challenging, may result into worse local minimum. In order to
overcome all these issues, one may consider an alternative
reparametrization $p_{ij}=\sigma(\bm{\hat{m}}_{ij})$, with $\bm{\hat{m}}_{ij}$
being a latent mask variable and $\sigma$ the sigmoid function which bounds
$p_{ij}$ in $[0,1]$. However, this workaround suffers (in practice) from
numerical instability in gradient estimation (due to the log and the sigmoid)
and is also computationally demanding. \\

\noindent\textbf{Arbitrarily Shifted Log Parametrization.}
Another alternative is to consider $\bm{\hat{m}}_{ij} =
\log(p_{ij})$ and $\log(1-p_{ij}) = \log(1-\exp(\bm{\hat{m}}_{ij}))$ and learn
the underlying mask. However, this reparametrization is also flawed in the
same way as the aforementioned sigmoid reparametrization. In what follows, we
propose an equivalent formulation which turns out to be highly effective and
numerically more stable. Considering  

\begin{equation}
  \begin{bmatrix}
    \bm{\hat{m}}_{ij} \\
    0  \\
  \end{bmatrix}
  = \log\big(\mathcal{P}_{ij}(.)\big) + c =
  \begin{bmatrix}
    \log(p_{ij}) + c \\
    \log(1-p_{ij}) + c\\
  \end{bmatrix},
  \label{eq:our-formulation}
\end{equation}

\noindent in the above definition, instead of using $\log(\mathcal{P}_{ij}(.))$, we
consider $\log(\mathcal{P}_{ij}(.)) + c$  as an input of the argmax in
\cref{eq:gumbel-softmax-argmax}. The constant $c \in \mathds{R}$ ensures that
if $\bm{\hat{m}}_{ij} > 0$, then $\log(p_{ij}) \in ]-\infty,0] \Leftrightarrow
p_{ij} \in [0,1]$. This is enforced by setting the second coefficient of
$\mathcal{P}_{ij}$ to 0, rather than computing it explicitly. The formulation
of \cref{eq:our-formulation} is theoretically equivalent to the aforementioned
sigmoid reparametrization. Indeed, solving the system of
\cref{eq:our-formulation} w.r.t. $\bm{\hat{m}}_{ij}$ yields $p_{ij} =
\sigma(\bm{\hat{m}}_{ij})$. \noindent Differently put, the formulation in
\cref{eq:our-formulation} considers a reparametrization $\bm{\hat{m}}_{ij} =
\log(p_{ij})+c$ which is strictly equivalent to the sigmoid one while being
computationally more efficient and also stable. Note that adding any arbitrary constant $c$ to the log-probability
makes the outcome of Gumbel-Softmax sampling and argmax invariant.

  \subsection{Weight Rescaling}
  \label{sec:smart-rescale}
Subnetwork selection may disrupt the dynamic of the forward pass
\cite{DBLP:conf/iccv/HeZRS15,DBLP:conf/cvpr/RamanujanWKFR20}, and thereby
requires adapting  weights accordingly.    Dynamic weight rescale (DWR)
\cite{DBLP:conf/nips/ZhouLLY19}, and scaled Kaiming distribution
\cite{DBLP:conf/cvpr/RamanujanWKFR20}  are two  known mechanisms that adapt the
weights of the selected subnetworks.  However, some of these heuristics, besides
being handcrafted,   rely on the strong assumption that rescaling should be
proportional to the pruning rate.  In what follows,  we consider a new weight
adaptation mechanism, referred to as Smart Rescale (SR). Instead of handcrafting
this rescaling factor proportionally to the pruning rate (as achieved for
instance in \cite{DBLP:conf/nips/ZhouLLY19}), SR is learned layerwise and
provides an effective (and also efficient)  way to adapt the dynamic of the
forward pass without retraining the entire weights of the selected subnetwork.
Indeed, this rescaling ends up reducing the amount of epochs needed to reach
convergence and also improving accuracy (at some extent) as shown later in
experiments. \\ With SR, the $\ell$-th layer network output becomes 
  \begin{equation}
   \mathbf{z}_{\ell} = g_\ell(s_\ell \times (\bm{m}_\ell \odot \bm{w}_\ell) \otimes \mathbf{z}_{\ell-1}),
  \end{equation}
  \noindent where  $s_\ell$ refers to the rescaling factor  of  the $\ell$-th
  layer (see also algorithm~\ref{alg:gumbel-forward}).  Smart Rescale increases
  the flexibility of subnetwork selection and adaptation compared to DWR (which
  is bound to the pruning rate).  Moreover,  scaling factors obtained with SR
  vary smoothly  ---  and this makes training more stable with stochastic
  gradient descent (SGD) --- compared to the ones obtained with DWR which are
  again set to the observed {\it pruning rates},  and changes of the latter are
  more abrupt due to stochastic mask sampling.    
  \begin{algorithm}
    \caption{Forward pass for our method}
    \label{alg:gumbel-forward}
    \begin{algorithmic}[1]
    \Require A network $f_\theta$, with weights $\{\bm{w}_\ell\}_\ell$, ASLP  $\{\bm{\hat{m}}_\ell\}_\ell$, and input training data $\{({\bf x}_k,{\bf y}_k)\}_k$ 
    \State
    $q_{i,j} \gets \text{argmax} 
    \begin{bmatrix}
      \bm{\hat{m}}_{i,j} + g_{i,j} \\
      0 + g'_{i,j}\\
    \end{bmatrix}$ \Comment{Sampling of a topology} 
    \State $m_{ij} \gets 1_{\{q_{ij}=1\}}$
   \Comment{
    Giving the masks $\bm{m}_{i,j}$ their values} 
    \State {\bf Return} ${\cal L}\big(f_\theta(\{{\bf x}_k\}_k; \{s_\ell (\bm{m}_\ell \odot \bm{w}_\ell)\}_\ell),\{{\bf y}_k\}_k\big)$
    \Comment{Computing the loss with masked weights and SR}
    \end{algorithmic}
  \end{algorithm} 
  \section{Experiments}\label{sec:experiments}
  
  In this section, we show the performance of our method on the standard CIFAR10
  and CIFAR100 datasets.   They consist of  60k  colored images of $32\times 32$
  pixels each.  Training,  validation and test sets  include  45k,  5k and 10k
  images respectively.  \\  In order to demonstrate the effectiveness of our
  method, we chose the widely used SGD optimizer with a momentum of  0.9  and a
  learning rate of 50.    Faster convergence is obtained with higher learning
  rates,   however,  the latter also lead  to worse observed accuracy.   During
  training,  the maximum number of epochs is set to 1000 and early stopping is
  triggered  if the accuracy on the validation set stops improving during 100
  epochs.   In all these experiments, neither weight decay nor $\ell_2$
  regularization are applied. See implementation details and our code on the ASLP GitHub
  \cite{Dupont2022}.

\begin{table*}[htbp]
  \centering
  \resizebox{14.0cm}{!} 
{
  \begin{tabular}{@{}llcccccccc|c@{}}
    \cline{3-11}
                       &                                                        & \multicolumn{8}{c|}{Cifar 10}                                                                                                 & Cifar 100 \\ 
                       \cline{3-11}
                            &                                                       & \multicolumn{4}{c}{w/o data augmentation}                                     & \multicolumn{4}{c|}{with data augmentation (w.d.a)}         &  w.d.a  \\
                            &                                                        & $\varnothing$ & WR             & SC            & WR+SC          & $\varnothing$ & WR             & SC            & WR+SC          &           WR+SC   \\ \midrule
    \multirow{4}{*}{Conv2} & \cite{DBLP:conf/nips/ZhouLLY19} (averaging)         &64.4          & 65.0          & 66.3          & 66.0          & -             & -             & -             & -            &               -        \\
                            & \cite{DBLP:conf/cvpr/RamanujanWKFR20}\footnotemark ($k=50\%$)  &               &   -           &         -     &         -     &       -       &      -        & 71.5   &       71.7       &     40.9         \\ 
                            & Our ASLP (averaging)                                     & 68.2          & 66.9          & 68.3          & 66.5          & \textbf{76.0} & \textbf{76.6} & 76.8          & 77.3          &       -         \\
                            & Our ASLP (thresholding)                                     & \textbf{68.7} & \textbf{67.8} & \textbf{68.4} & \textbf{67.1} & 75.9          & 76.4          & \textbf{77.5} & \textbf{77.5} &    \textbf{43.3}          \\
                            \midrule
  
    \multirow{4}{*}{Conv4} & \cite{DBLP:conf/nips/ZhouLLY19} (averaging)       & 65.4          & 71.1          & 66.2          & 72.5          & -             & -             & -             & -             &             -           \\
                            & \cite{DBLP:conf/cvpr/RamanujanWKFR20}\footnotemark[\value{footnote}] ($k=50\%$) &        -      &  -            &            -  &  -            &  -            &  -            & 81.6  &     80.5    &  51.1  \\ 
                            & Our ASLP (averaging)                                          & 70.6          & 71.8          & 69.5          & 71.8          & 83.4          & 84.4          & 83.7          & 84.1          &        -   \\
                            & Our ASLP (thresholding)                                      & \textbf{71.5}  & \textbf{72.8}& \textbf{70.2} & \textbf{72.7} & \textbf{83.7}& \textbf{85.0} & \textbf{84.5} & \textbf{84.8} &         \textbf{51.7}  \\
                            \midrule
    \multirow{4}{*}{Conv6} & \cite{DBLP:conf/nips/ZhouLLY19} (averaging)    & 63.5          & 76.3          & 65.4          & 76.5          & -             & -             & -             & -      &  -       \\
                            & \cite{DBLP:conf/cvpr/RamanujanWKFR20}\footnotemark[\value{footnote}] ($k=50\%$) &      -        &    -          &        -      &     -         &      -        &           -   &  85.4 &    85.1   &   \textbf{53.8}   \\ 
                            & Our ASLP (averaging)                                      & 72.9          & 76.1          & 71.9          & 75.6          & 85.3          & 86.2          & 85.3          & 86.2          &  - \\
                            & Our ASLP (thresholding)                                       & \textbf{73.7} &\textbf{77.0}  &\textbf{72.6}  &\textbf{76.6}  &\textbf{86.0}  &\textbf{86.9}  & \textbf{86.3} &\textbf{86.9}  &   52.8 \\
                            \bottomrule
    \end{tabular}
}

  \caption{\footnotesize Comparison of our method against \cite{DBLP:conf/nips/ZhouLLY19}
  and \cite{DBLP:conf/cvpr/RamanujanWKFR20} on Conv2, Conv4 and Conv6. These
  results are averaged through five independent runs. "WR" (Weight Rescale)
  refers to ``Dynamic Weight Rescale'' or ``Smart Rescale'' depending on which
  methods is used (respectively \cite{DBLP:conf/nips/ZhouLLY19} or our proposed
  ASLP). Again, "SC" refers to the ``Signed Constant'' distribution. The latest results on CIFAR 100
  were recently obtained with data augmentation and WR+SC.}
  \label{tbl:conv_compare}

\end{table*}

\subsection{Performance and comparison}
The accuracy of our method is evaluated on subnetworks whose topology
corresponds to connections with  (trained) probabilities larger than 0.5;  in
other words,  if {\it the binary event of keeping a connection is more likely
than its removal}.  This setting is referred to as  {\it thresholding}.  As a
matter of comparison,  we also consider the setting in
\cite{DBLP:conf/nips/ZhouLLY19} which consists in sampling ten different
subnetworks and evaluating an average accuracy over  these subnetworks.  This
setting  is referred to as {\it averaging}.  In these experiments,  we use the
same networks as
\cite{DBLP:conf/nips/ZhouLLY19,DBLP:conf/cvpr/RamanujanWKFR20} (originally
introduced by  Frankle and Carbin \cite{DBLP:conf/iclr/FrankleC19}) namely
Conv2,  Conv4 and Conv6 which are  variants of  VGG16. \\
\indent   \cref{tbl:conv_compare}  shows a comparison of our method against
\cite{DBLP:conf/nips/ZhouLLY19,DBLP:conf/cvpr/RamanujanWKFR20}.    These
results show means of five independent runs; each run corresponds either to
``thresholding'' or ``averaging''.  These performances show a consistent gain
(in accuracy) of our subnetwork selection.   We also observe that
``thresholding'' is already effective compared to ``averaging''; indeed, our
method reaches a  high accuracy despite learning a single subnetwork topology,
and this makes it also highly efficient for training compared to  the related
work \cite{DBLP:conf/nips/ZhouLLY19,DBLP:conf/cvpr/RamanujanWKFR20}.\\ 
\indent Furthermore, our method and \cite{DBLP:conf/nips/ZhouLLY19} do not
impose a pruning rate. The optimal pruning rate is found during optimization and
is is arround 51\%, whereas \cite{DBLP:conf/cvpr/RamanujanWKFR20} enforces a 50\%
pruning rate ($k=50\%$). Thus, the networks capacities
are comparable. \\

  \subsection{Ablation study}
  In this section, we discuss the impact of all the components of the method
  when taken individually and combined,  namely the use of weight rescaling
  (WR): either DWR or our proposed SR.   We also consider another criterion:
  signed constant  (SC) which consists in replacing weights in a given layer by
  the products of their signs and the standard deviation of their original
  weight distribution.  We show  all these results with and without data
  augmentation, which is composed of the combination of zero-padding,  random crops and
  random  horizontal flips.   Note that  pixel intensities are normalized  from
  their original values in $[0,255]$ to $[0,1]$.
    
  From the results in table~\ref{tbl:conv_compare}, we observe a clear gain of
  our method alone w.r.t.  \cite{DBLP:conf/nips/ZhouLLY19} and the use of SR
  increases further its accuracy (excepting Conv2 w/o data augmentation).  The
  gain in performances increases  significantly with Conv6 and reaches up to 4
  points even when no data augmentation is used.   Note that the use of data
  augmentation attenuates, at some extent, the effect of SR on larger networks
  (conv4 and Conv6). Nonetheless,  as discussed in \cref{sec:SR-impact}, the
  positive impact of SR resides also in training efficiency.   In contrast to
  SR, signed constant improves accuracy by a small margin when combined with
  data augmentation. 
  
  \footnotetext{Performances for \cite{DBLP:conf/cvpr/RamanujanWKFR20} are reported
  with the optimizer described in \cref{sec:experiments}. It is possible to
  improve performances by tuning the learning rate scheduler but this is out of
  the scope of this paper.}
  
  \subsection{Computational efficiency}
  \label{sec:SR-impact}
DWR requires rectifying weights layerwise using the inverse of the observed
(computed) pruning rates. These layerwise evaluations introduce a significant
overhead at each training epoch. In contrast, SR consists in simple products
involving one scalar per layer. When training Conv4, we found (on average) that
enabling DWR increases epoch runtime by $0.2s$ while our SR by $0.13s$ only, so
SR speeds up training overhead by 35\% compared to DWR.  When data augmentation
and signed constant are used, SR allows a significant gain in the number of
training epochs. Indeed, enabling SR on Conv4  saves (on average) 19.7\%
training epochs (8.2\%, 14.0\%  on Conv2 and Conv6 respectively) before
converging to its highest accuracy. Finally, our ``thresholding'' setting not
only improves accuracy but makes subnetwork selection (training) and also
inference more efficient compared to the related work
\cite{DBLP:conf/nips/ZhouLLY19,DBLP:conf/cvpr/RamanujanWKFR20},  as this
selection is again achieved  once and thereby only one subnetwork is applied
during inference.

\section{Conclusion}

In this paper, we introduce a novel method that extracts effective subnetworks
from larger networks without training its weights. The proposed method optimizes
a probability distribution which measures the relevance of weights, and only
those with the highest relevance define the topology of the selected
subnetworks. An efficient and effective weight rescaling mechanism is also
introduced and allows rectifying the parameters of the selected subnetworks
which improves performances and reduces the number epochs needed to reach
convergence. Experiments conducted on the standard CIFAR10 and  CIFAR100
datasets show the effectiveness of our subnetwork selection method w.r.t. the
related work. Future work includes the study of the scalability of the proposed
method on more complex datasets and other larger networks.\\

\noindent\textbf{Acknowledgement.} 
This work was performed using HPC resources from GENCI-IDRIS (Grant 2021-AD011011427R1). \\
It has been achieved within a partnership between Sorbonne University and Netatmo.

\noindent\textbf{Code.} Our code is available at:\\ \href{https://github.com/N0ciple/ASLP}{\texttt{https://github.com/N0ciple/ASLP}}


\bibliographystyle{IEEEbib}
\bibliography{dupont}

\end{document}